\documentclass[letterpaper]{article} 
\usepackage{aaai2026}  
\usepackage{times}  
\usepackage{helvet}  
\usepackage{courier}  
\usepackage[hyphens]{url}  
\usepackage{graphicx} 
\usepackage{natbib}  
\usepackage{caption} 
\usepackage{algorithm}
\usepackage{algorithmic}

\usepackage{newfloat}
\usepackage{listings}
\DeclareCaptionStyle{ruled}{labelfont=normalfont,labelsep=colon,strut=off} 
\floatstyle{ruled}
\newfloat{listing}{tb}{lst}{}
\floatname{listing}{Listing}
\title{Search-Based Credit Assignment for Offline Preference-Based Reinforcement~Learning}

\author {
    Xiancheng Gao\textsuperscript{\rm 1},
    Yufeng Shi\textsuperscript{\rm 1},
    Wengang Zhou\textsuperscript{\rm 1,2},
    Houqiang Li\textsuperscript{\rm 1,2}
}
\affiliations {
    \textsuperscript{\rm 1}EEIS Department, University of Science and Technology of China \\
    \textsuperscript{\rm 2}Institute of Artificial Intelligence, Hefei Comprehensive National Science Center, Hefei, China \\
    \{gxcnb, shiyufeng\}@mail.ustc.edu.cn, \{zhwg, lihq\}@ustc.edu.cn
}

\nocopyright

\usepackage{booktabs}
\usepackage{multirow}
\usepackage[table]{xcolor}   
\usepackage{bm}
\usepackage{amsfonts}
\usepackage{amsmath}
\definecolor{best}{HTML}{FFEFC6}
\definecolor{second}{gray}{0.9}
\newcommand{\secondcell}[1]{\cellcolor{second}{#1}}
\newcommand{\bestcell}[1]{\cellcolor{best}{#1}} 

\begin{document}

\maketitle

\begin{abstract}
Offline reinforcement learning refers to the process of learning policies from fixed datasets, without requiring additional environment interaction. However, it often relies on well-defined reward functions, which are difficult and expensive to design.  Human feedback is an appealing alternative, but its two common forms, expert demonstrations and preferences, have complementary limitations. Demonstrations provide stepwise supervision, but they are costly to collect and often reflect limited expert behavior modes. In contrast, preferences are easier to collect, but it is unclear which parts of a behavior contribute most to a trajectory segment, leaving credit assignment unresolved. 
In this paper, we introduce a Search-Based Preference Weighting (SPW) scheme to unify these two feedback sources. 
For each transition in a preference labeled trajectory, SPW searches for the most similar state–action pairs from expert demonstrations and directly derives stepwise importance weights based on their similarity scores. 
These weights are then used to guide standard preference learning, enabling more accurate credit assignment that traditional approaches struggle to achieve. 
We demonstrate that SPW enables effective joint learning from preferences and demonstrations, outperforming prior methods that leverage both feedback types on challenging robot manipulation tasks.
\end{abstract}


\begin{figure*}[t]
    \centering
    \includegraphics[width=0.9\linewidth]{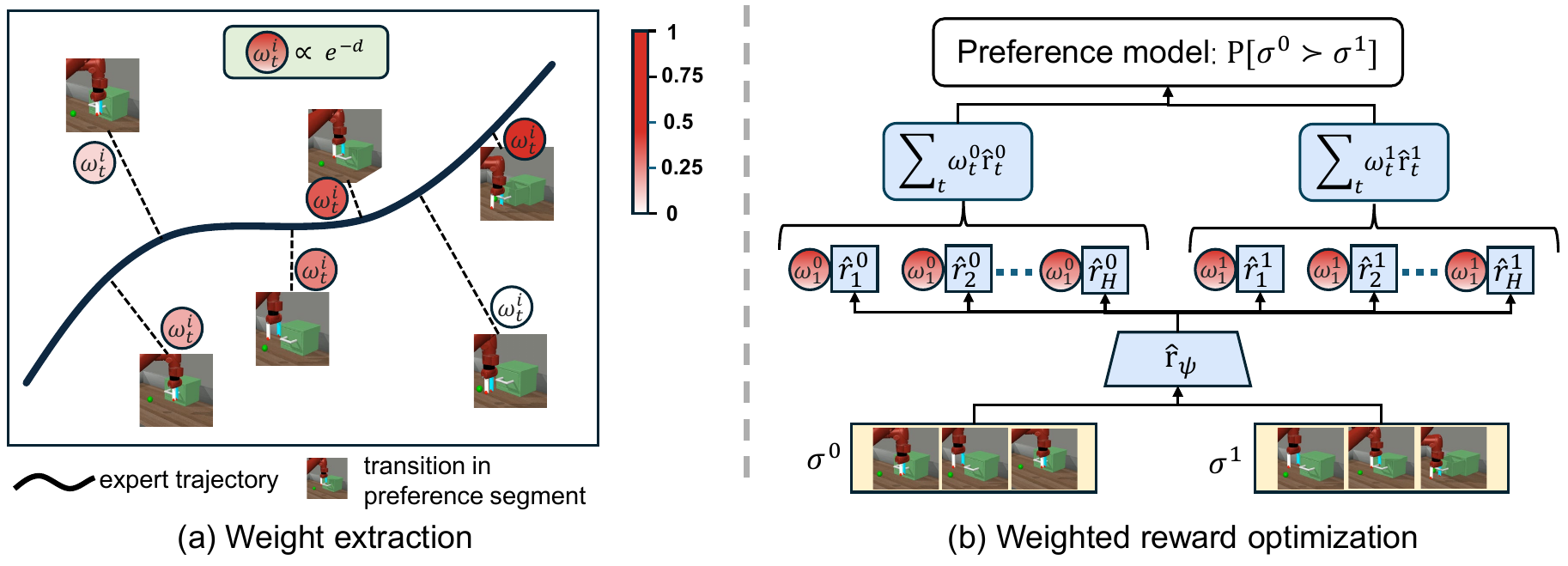}
    \caption{\textbf{An overview of SPW.}
(a) Weight extraction: Importance weights are computed for each transition in the preference-labeled trajectories based on their similarity to expert demonstrations. Darker colors indicate higher weights.
(b) Weighted reward optimization: The return of a preference trajectory is modeled as the weighted sum of stepwise rewards. This formulation is then integrated into the standard preference learning framework.}
    \label{fig:overview}
\end{figure*}

\section{Introduction}
Reinforcement learning (RL) is a paradigm in machine learning where an agent learns to make sequential decisions by interacting with an environment and being guided by reward signals.
RL has recently excelled in video games~\citep{mnih2015dqn}, robotic manipulation~\citep{kalashnikov2018qtop}, nuclear‑fusion control~\citep{degrave2022fusion}, and autonomous driving~\citep{kiran2021survey}. These successes, however, hinge on carefully engineered reward functions that are costly to design and often misaligned with human intent. 

To bypass this bottleneck, researchers learn rewards directly from human feedback, chiefly expert demonstrations and trajectory‑level preferences.
Expert demonstrations offer detailed step‐by‐step instructions on completing a task and serve as a dense source of supervision. However, they are costly to collect and often limited in behavioral diversity. 
In contrast, trajectory preferences are easy to collect since humans only need to choose the better of two trajectories. However, they offer only coarse‑grained supervision and lack transition‑level indication to identify which states or actions led to the preference. 
This raises the key challenge in preference‑based reward learning: how to assign credit when only overall trajectory comparisons are given.

Although expert demonstrations and trajectory‑level preferences have their own strengths, neither of them is sufficient on its own for robust, human‑aligned reward learning. 
Demonstrations convey fine‑grained structure, whereas preferences reflect broader judgments about overall task progress. 
Based on this observation, some researchers try to fuse them to take the advantage of them. 
For example, in~\cite{ibarz2018reward}, a policy is initialized with behavior cloning~\cite{pomerleau1988alvinn} and then fine‑tuned using preference supervision. 
In~\cite{byk2021bayesian}, a Bayesian framework is introduced to build a prior from demonstrations and repeatedly refines it with preference queries.
These pipelines, however, treat demonstrations and preferences in separate stages, limiting synergy and overall performance. 
More recently, in~\cite{taranovic2023ailp}, the two signals are unified in an adversarial imitation learning setting, but this method requires online interaction and suffers from interference between imitation and preference losses, which destabilizes training.

Our method takes a different perspective from the above approaches. 
Specifically, we explicitly examine the strengths and shortcomings of imitation learning and preference-based approaches. 
This helps identify how they can complement each other within a unified framework. Classical imitation learning methods \citep{pomerleau1988alvinn,ho2016gail,kostrikov2020sqil} and newer similarity‑based approaches \citep{dadashi2021primal,luo2023optimal,lyu2024seabo} generally assume uniform importance across expert transitions. As a result, they fail to capture task progress and goal completion \citep{liu2023imitation}. Preference‑based reward learning (PbRL) algorithms \citep{christiano2017deep} offer an attractive alternative. However, most rely on the BT model \citep{bradley1952rank} and leave within-trajectory credit assignment unaddressed.

Motivated by these gaps, we propose a \textbf{S}earch‑Based \textbf{P}reference \textbf{W}eighting (\textbf{SPW}) scheme, which unifies preference and demonstration supervision without extra loss terms, multistage optimization, or online interaction. 
Specifically, SPW first matches each transition in a preference trajectory with the most similar state–action pairs from expert demonstrations. 
It then derives stepwise importance weights from these matches. 
These weights refine the BT model for fine‑grained credit assignment, tackling the key limitation of PbRL.
Since trajectory comparisons naturally reveal task progress and goal relevance, SPW also informs how demonstrations are weighted. 
This relaxes the uniform‑importance assumption in imitation learning as a byproduct.

Our main contributions are summarized as follows.
\begin{itemize}
    \item  We provide the first analysis of the credit assignment challenge in PbRL, clarifying why existing approaches struggle to determine which states or actions drive preferences.
    
    \item We introduce SPW, a lightweight method that integrates demonstrations and preferences into a single-stage framework without requiring additional loss terms or online interaction. It effectively addresses credit assignment in PbRL.

    \item  On standard robotic manipulation benchmarks, SPW improves success rates with only a small number of demonstrations and a few hundred preference labels, outperforming imitation, preference-based, and hybrid baselines in most cases.
\end{itemize}

\section{Related Work}
\subsection{Learning from Preferences}
Preference-based reinforcement learning (PbRL) infers reward functions from pairwise trajectory comparisons. Recent works aim to reduce the amount of human feedback required in PbRL, such as using unsupervised exploration for policy pretraining \citep{lee2021pebble}, meta-learning for generalization from few annotated pairs \citep{xiao2022meta}, semi-supervised learning and data augmentation to amplify preference data \citep{xu2022semi}, denoising discriminators for noisy labels \citep{meng2023denoising}, and suboptimal demonstrations for reward model pretraining \citep{muslimani2025demo}. The offline setting of PbRL introduces additional challenges by removing the possibility of interactive querying. To address this, recent methods avoid explicit reward modeling by directly optimizing policies from preferences \citep{tang2023direct,tian2023policy}, estimate listwise rewards from trajectory sets \citep{choi2024listwise}, model preference sequences using transformers \citep{kim2023preference}, or generate improved trajectories via diffusion models conditioned on preferences \citep{li2023diffusion,chen2024diffusionpref}. However, existing methods offer little analysis or resolution of the credit assignment problem.

\subsection{Learning from Demonstrations and Preferences}
Several recent works attempt to combine demonstrations and preferences to better leverage the complementary strengths of both forms of supervision. Previous research has provided theoretical insights into learning from ranked demonstrations \citep{jeon2020theory}, forming the basis for methods like T-REX \citep{brown2019trex}. D-REX \citep{brown2019drex} extends this idea by perturbing demonstration trajectories to synthesize ranking information. Other work demonstrates that policies can first be initialized via behavior cloning and then refined through PbRL \citep{ibarz2018reward}. Similarly, Bayesian methods start with an expert-driven reward prior and update it with human preferences \citep{biyik2021bayesian}, though both approaches require multiple learning stages and lead to a more complex training process.  More recently, AILP \citep{taranovic2023ailp} integrates demonstrations and preferences into an adversarial imitation learning framework. However, it jointly optimizes a reward model with both the discriminator loss from GAIL and preference loss, which may lead to conflicting optimization signals.

\section{Preliminaries}
We model the interaction between an agent and its environment as a Markov Decision Process (MDP), defined by the tuple $\langle \mathcal{S}, \mathcal{A}, p, r, \gamma, p_0 \rangle$, where $\mathcal{S}$ is the state space, $\mathcal{A}$ is the action space, $p: \mathcal{S} \times \mathcal{A} \times \mathcal{S} \rightarrow [0,1] $  denotes the transition probability function, $r: \mathcal{S} \times \mathcal{A} \rightarrow \mathbb{R}$ is the scalar reward function, $\gamma \in [0, 1)$ is the discount factor, and $p_0$ is the distribution over initial states. A policy $\pi(a|s)$ defines a distribution over actions given a state $s$. 
The objective in reinforcement learning is to find a policy that maximizes the expected cumulative discount reward:
\begin{equation}
J(\pi) = \mathbb{E}_{s_0 \sim p_0,\; a_t \sim \pi,\; s_{t+1} \sim p} \left[ \sum_{t=0}^{\infty} \gamma^t r(s_t, a_t) \right].
\label{eq:rl_objective}
\end{equation}

We consider a learning setting with a small number of expert demonstrations and a large collection of preference pairs. Let \( \mathcal{D}_e = \{ \xi_i \}_{i=1}^{N_e} \) denote the set of expert demonstrations, where each segment \( \xi = \{(s_1^e, a_1^e), \ldots, (s_T^e, a_T^e)\} \) represents a successful execution of the task. Note that the trajectory length \( T \) may vary between demonstrations.
We define \( \mathcal{E} \) as the set of all transitions in \( \mathcal{D}_e \), that is,
$\mathcal{E} = \bigcup_{i=1}^{N_e} \{ (s_t^{e_i}, a_t^{e_i}) \}_{t=1}^{T_i}$, which will be used for the similarity-based calculation.

In contrast, preference segments are collected from an unknown behavior policy $\mu$, and are more abundant and potentially suboptimal. Each segment is denoted as $\sigma = \{(s_1, a_1), \dots, (s_H, a_H)\}$.
Pairs of these segments are presented to a human annotator who is asked to indicate which is preferred. Given a pair $(\sigma^0, \sigma^1)$, a preference label $l$ is provided as $l = 0$ if $\sigma^0 \succ \sigma^1$, $l = 1$ if $\sigma^0 \prec \sigma^1$, and $l = 0.5$ if $\sigma^0 \sim \sigma^1$, where $\sigma^0 \succ \sigma^1$ indicates that $\sigma^0$ is preferred over $\sigma^1$, and $\sigma^0 \sim \sigma^1$ denotes indifference.

To learn a reward function $r_\psi$ from preference pairs, most methods adopt the BT model~\cite{bradley1952rank}, a fundamental paradigm in preference learning. This model estimates the probability that one trajectory segment is preferred over another on the basis of their cumulative returns. Specifically, given a pair of trajectory segments $\sigma^0$ and $\sigma^1$, the probability of preference is defined as:

\begin{equation}
P_{\psi}[\sigma^0 \succ \sigma^1] = 
\frac{
\exp\left( \sum_t \hat{r}_{\psi}(s_t^0, a_t^0) \right)
}{
\sum_{i=0}^1 \exp\left( \sum_t \hat{r}_{\psi}(s_t^i, a_t^i) \right)
}.
\label{eq:bt}
\end{equation}

Then given a preference dataset, the reward function $\hat{r}_{\psi}$ is updated by minimizing the cross-entropy between $P_{\psi}$ and the ground-truth preference labels:
\begin{align}
\mathcal{L}_{\text{CE}} = -\mathbb{E}_{(\sigma^0, \sigma^1, l) \sim \mathcal{D}_{pref}} \Big[
& (1-l) \log P_{\psi}[\sigma^0 \succ \sigma^1] \notag \\
& +\; l \log P_{\psi}[\sigma^1 \succ \sigma^0]
\Big].
\label{eq:ce_loss}
\end{align}

\begin{figure}[t]
    \centering
    \includegraphics[width=\linewidth]{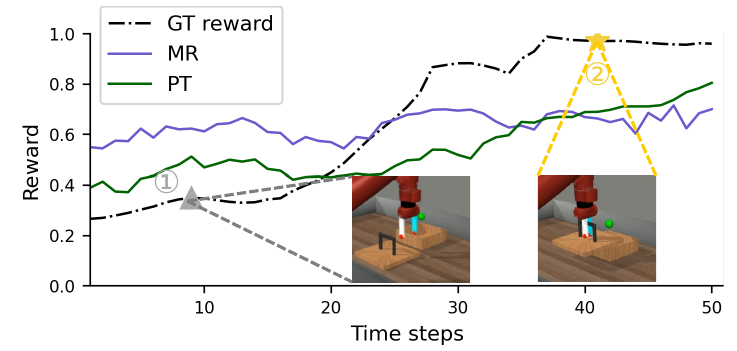}
    \caption{Normalized reward curves of MR, PT, and GT for trajectory segments in the \textit{box-close} task. Snapshots from selected positions along the segment are shown for visual reference.}
    \label{fig:method_credit}
\end{figure}

\section{Method}
\subsection{Motivation}
In PbRL, the feedback is sparse. 
For a pair of trajectories, we only observe which trajectory is preferred, without information about the reasons for that preference. 
Consequently, it is a challenge to assign credit to the individual transitions that gave rise to the preference. 
Psychological evidence also suggests that human judgments often depend on a small number of decisive moments~\citep{Kahneman2000}. 
This reinforces the need for fine‑grained credit assignment.

Most PbRL algorithms adopt the standard BT model to optimize the reward function, as shown in Eq.~\ref{eq:bt}.
However, our investigation shows that this approach struggles with credit assignment. 
As shown in Figure~\ref{fig:method_credit}, the rewards predicted by MR, a standard offline PbRL algorithm, within each trajectory segment are almost flat.
These predicted rewards fail to highlight the transitions that matter and differ significantly from the ground-truth reward (GT).

This discrepancy is rooted in the learning objectives of the BT model. 
A BT probability only needs to make the preferred trajectory score slightly higher than the inferior one. 
Once the aggregate margin
$\Delta = \sum_t \hat r_\psi(\sigma^{0}) - \sum_t \hat r_\psi(\sigma^{1})$
turns positive, the cross‑entropy loss stops pushing the two trajectories apart. 
As a result, most states converge to similar “average” rewards. 
The learned reward distribution then centers in the middle. Further analysis is provided in the subsequent experiments section.

Recent work, Preference Transformer (PT)~\citep{kim2023preference}, attempts to address this issue by replacing the MLP-based reward model with a transformer.
However, PT still relies solely on trajectory‑level labels, which makes its attention weights difficult to interpret. 
As shown in Figure~\ref{fig:method_credit}, PT produces a more differentiated reward curve, but it still deviates substantially from the GT reward.

This analysis illustrates the limitations of existing approaches and underscores the need for a mechanism that can attribute credit to the most influential transitions rather than treating all steps equally. 

To address this, we propose SPW, which assigns transition-level importance weights by comparing preference trajectories to expert behavior. These weights guide the reward model to focus on key transitions, enabling finer-grained credit assignment and more effective learning from coarse preference labels.

\subsection{Expert-Guided Importance Weighting}  
Human preferences over trajectory segments are often modeled as the unweighted sum of per‑step rewards, assuming all transitions contribute equally.  
In reality, some transitions matter more.  
SPW captures this by computing transition-level weights through comparison with expert behavior.

\subsubsection{Nearest‑Neighbor Distance.}  
Let $\mathcal{E}$ denote the set of expert transitions.  
For each transition $(s,a)$ in a preference segment, we measure its distance to the closest expert transition using a KD-tree~\citep{bentley1975multidimensional} and a chosen metric \( D \) (e.g., Euclidean distance):  
\begin{equation}
d = D\bigl((s,a), (s^\ast, a^\ast)\bigr),
\label{eq:distance}
\end{equation}
where $(s^\ast, a^\ast) \in \mathcal{E}$ is the nearest neighbor.  
A smaller distance indicates closer alignment with expert behavior, while a larger distance suggests deviation.

\subsubsection{Intra-Segment Weight Extraction.}  
Given a segment \( \sigma = \{(s_1, a_1), \dots, (s_H, a_H)\} \) with distances \( \{d_t\}_{t=1}^H \), we assign higher weights to transitions that better match expert behavior.  
Weights are computed via softmax normalization:
\begin{equation}
w_t = \frac{\exp(-d_t / \tau)}{\sum_{t'=1}^H \exp(-d_{t'} / \tau)},
\label{eq:softmax_weight}
\end{equation}
where \( \tau \) is a temperature hyperparameter. As \( \tau \to 0 \), the weight mass concentrates on the transition most similar to expert behavior; as \( \tau \to \infty \), the weights approach uniformity.  
This normalization ensures consistent reward magnitudes across segments, preventing bias in preference loss and stabilizing training.

\subsection{Weighted Preference Optimization}  
With these expert-guided weights, SPW reformulates human preferences as a weighted sum of per-step rewards rather than an unweighted sum. 
More influential transitions exert greater impact on the preference signal, enabling interpretable credit assignment. This perspective also aligns with cognitive studies suggesting that humans focus on a few salient moments when evaluating experiences \citep{Kahneman2000}.

Formally, the weighted return for a segment is:
\begin{equation}
R_{\sigma} = \sum_{t=1}^H w_t\, \hat{r}_{\psi}(s_t, a_t),
\label{eq:weighted_return}
\end{equation}
where \( w_t \) is the importance weight for transition \( t \) and \( \hat r_{\psi} \) is the learned reward model.

We integrate this weighted return into the Bradley–Terry (BT) preference model.  
For a labeled pair \( (\sigma^0, \sigma^1, l) \), the probability that segment \( \sigma^0 \) is preferred over \( \sigma^1 \) becomes:
\begin{equation}
P_{\psi}\!\bigl[\sigma^0\!\succ\!\sigma^1\bigr] =
\frac{\exp\!\bigl(\sum_{t} w_t^{0}\,\hat r_{\psi}(s_t^{0},a_t^{0})\bigr)}
     {\sum_{i=0}^{1} \exp\!\bigl(\sum_{t} w_t^{i}\,\hat r_{\psi}(s_t^{i},a_t^{i})\bigr)}.
\label{eq:weighted_bt}
\end{equation}

The reward model \( \hat r_{\psi} \) is trained by maximizing the log-likelihood of labeled preference pairs (Eq.~\ref{eq:ce_loss}).  
By incorporating weights \( w_t^0, w_t^1 \), the model can focus on key transitions that align with expert behavior, leading to more fine-grained credit assignment and more effective learning from coarse preference labels.

\begin{table*}[t]
  \centering
  \footnotesize
  \setlength{\tabcolsep}{4pt} 
  \begin{tabular}{c|l|*{8}{c}}
    \toprule
    \textbf{Feedbacks} & \textbf{Algorithm} &
    peg‑unplug & box‑close & sweep & sweep‑into &
    handle‑pull & dial‑turn & drawer‑open & lever‑pull \\
    \midrule \midrule

    \multirow{6}{*}{100} &
    MR &
    32.8 $\pm$ {\tiny 14.42} & 1.6 $\pm$ {\tiny 3.58} &
    40.4 $\pm$ {\tiny 28.51} & 32.4 $\pm$ {\tiny 14.26} &
    30.0 $\pm$ {\tiny 13.87} & 34.0 $\pm$ {\tiny 14.75} &
    \secondcell{92.4 $\pm$ {\tiny 8.08}} & 98.0 $\pm$ {\tiny 2.00} \\
    
    & OPRL  &
    32.0 $\pm$ {\tiny 12.76} & 0.0 $\pm$ {\tiny 0.00} &
    31.2 $\pm$ {\tiny 12.70} & 33.6 $\pm$ {\tiny 12.29} &
    39.6 $\pm$ {\tiny 5.02} & 42.4 $\pm$ {\tiny 13.31} &
    89.2 $\pm$ {\tiny 10.06} & \bestcell{98.8 $\pm$ {\tiny 12.47}} \\
    
    & PT &
    32.4 $\pm$ {\tiny 13.73} & 0.8 $\pm$ {\tiny 1.79} &
    0.0 $\pm$ {\tiny 0.00} & 45.6 $\pm$ {\tiny 21.80} &
    21.6 $\pm$ {\tiny 12.62} & 26.4 $\pm$ {\tiny 11.87} &
    72.4 $\pm$ {\tiny 13.09} & 75.6 $\pm$ {\tiny 11.70} \\

    & IPL  &
    25.4 $\pm$ {\tiny 15.72} & 4.2 $\pm$ {\tiny 3.12} &
    19.2 $\pm$ {\tiny 9.56} & 27.0 $\pm$ {\tiny 11.34} &
    1.8 $\pm$ {\tiny 2.24} & 33.4 $\pm$ {\tiny 12.28} &
    51.4 $\pm$ {\tiny 9.21} & 29.1 $\pm$ {\tiny 12.26} \\

    & LiRE  &
    \secondcell{34.8 $\pm$ {\tiny 23.22}} & \secondcell{16.8 $\pm$ {\tiny 20.47}} &
    \secondcell{46.4 $\pm$ {\tiny 13.82}} & \secondcell{54.8 $\pm$ {\tiny 13.16}} &
    \bestcell{61.2 $\pm$ {\tiny 5.22}} & \secondcell{44.8 $\pm$ {\tiny 12.95}} &
    80.0 $\pm$ {\tiny 13.79} & 94.8 $\pm$ {\tiny 4.82} \\

    \cmidrule(lr){2-10}
    & \textbf{SPW (ours)} &
    \bestcell{45.2 $\pm$ {\tiny 15.38}} & \bestcell{46.8 $\pm$ {\tiny 14.86}} &
    \bestcell{50.0 $\pm$ {\tiny 16.46}} & \bestcell{57.6 $\pm$ {\tiny 12.60}} &
    \secondcell{53.2 $\pm$ {\tiny 10.77}} & \bestcell{46.4 $\pm$ {\tiny 6.42}} &
    \bestcell{99.2 $\pm$ {\tiny 1.15}} & \secondcell{98.4 $\pm$ {\tiny 1.50}} \\
    \midrule

    \multirow{6}{*}{200} &
    MR &
    28.4 $\pm$ {\tiny 13.02} & 2.0 $\pm$ {\tiny 3.46} &
    23.2 $\pm$ {\tiny 13.23} & 24.8 $\pm$ {\tiny 10.95} &
    27.4 $\pm$ {\tiny 10.59} & 44.0 $\pm$ {\tiny 13.04} &
    \secondcell{94.8 $\pm$ {\tiny 6.57}} & \secondcell{95.2 $\pm$ {\tiny 3.63}} \\

    & OPRL &
    32.0 $\pm$ {\tiny 12.13} & 2.4 $\pm$ {\tiny 1.38} &
    52.8 $\pm$ {\tiny 11.71} & 47.6 $\pm$ {\tiny 13.49} &
    37.6 $\pm$ {\tiny 14.26} & 47.2 $\pm$ {\tiny 10.32} &
    88.0 $\pm$ {\tiny 7.22} & 95.8 $\pm$ {\tiny 11.75} \\

    & PT &
    32.4 $\pm$ {\tiny 14.11} & 0.8 $\pm$ {\tiny 1.79} &
    0.0 $\pm$ {\tiny 0.00} & 54.4 $\pm$ {\tiny 11.54} &
    24.8 $\pm$ {\tiny 3.46} & 43.2 $\pm$ {\tiny 13.38} &
    83.2 $\pm$ {\tiny 9.90} & 72.8 $\pm$ {\tiny 10.23} \\   
    
    & IPL &
    28.6 $\pm$ {\tiny 14.22} & 9.2 $\pm$ {\tiny 7.48} &
    36.4 $\pm$ {\tiny 10.85} & 25.6 $\pm$ {\tiny 10.04} &
    10.6 $\pm$ {\tiny 14.23} & 36.6 $\pm$ {\tiny 12.50} &
    52.6 $\pm$ {\tiny 13.33} & 36.2 $\pm$ {\tiny 16.53} \\

    & LiRE &
    \secondcell{39.6 $\pm$ {\tiny 22.86}} & \secondcell{25.6 $\pm$ {\tiny 17.53}} &
    28.4 $\pm$ {\tiny 13.34} & \secondcell{56.0 $\pm$ {\tiny 7.48}} &
    \bestcell{65.4 $\pm$ {\tiny 5.18}} & \bestcell{50.4 $\pm$ {\tiny 11.28}} &
    90.0 $\pm$ {\tiny 12.36} & 93.2 $\pm$ {\tiny 9.86} \\

    \cmidrule(lr){2-10}
    & \textbf{SPW (ours)} &
    \bestcell{63.2 $\pm$ {\tiny 14.60}} & \bestcell{41.2 $\pm$ {\tiny 15.45}} &
    \bestcell{74.8 $\pm$ {\tiny 11.64}} & \bestcell{64.4 $\pm$ {\tiny 16.52}} &
    \secondcell{59.0 $\pm$ {\tiny 6.00}} & \secondcell{48.4 $\pm$ {\tiny 5.90}} &
    \bestcell{100.0 $\pm$ {\tiny 0.00}} & \bestcell{98.4 $\pm$ {\tiny 2.61}} \\
    \bottomrule
  \end{tabular}
  \caption{Success rate (\%) of PbRL baselines in eight Meta‑World tasks with 100 and 200 preference feedbacks. 
  Results are reported as mean ± standard deviation across five seeds. In the table, “peg-unplug” denotes peg-unplug-side, and “handle-pull” denotes handle-pull-side.
  Yellow and gray shading indicate the best and second‑best performances, respectively.}
  \label{tab:pbrl}
\end{table*}

\begin{table*}[t]
  \centering
  \footnotesize
  \setlength{\tabcolsep}{4pt} 
  \begin{tabular}{c|l|*{8}{c}}
    \toprule
    \textbf{Feedbacks} & \textbf{Algorithm} &
    peg‑unplug & box‑close & sweep & sweep‑into &
    handle‑pull & dial‑turn & drawer‑open & lever‑pull \\
    \midrule \midrule

    \multirow{1}{*}{--} &
    BC &
    20.0 $\pm$ {\tiny 4.24} & 17.6 $\pm$ {\tiny 7.52} &
    16.8 $\pm$ {\tiny 8.60} & 7.2 $\pm$ {\tiny 2.78} &
    29.6 $\pm$ {\tiny 5.13} & 13.2 $\pm$ {\tiny 4.49} &
    72.0 $\pm$ {\tiny 6.63} & 15.6 $\pm$ {\tiny 5.61} \\

    \multirow{1}{*}{--} &
    SEABO &
    17.2 $\pm$ {\tiny 6.10} & 0.4 $\pm$ {\tiny 0.89} &
    54.4 $\pm$ {\tiny 22.60} & 34.0 $\pm$ {\tiny 9.70} &
    12.0 $\pm$ {\tiny 6.62} & 10.8 $\pm$ {\tiny 5.02} &
    1.6 $\pm$ {\tiny 1.67} & 0.0 $\pm$ {\tiny 0.00} \\
    \midrule

    \multirow{5}{*}{100} &
    MR &
    \secondcell{32.8 $\pm$ {\tiny 14.42}} &  \secondcell{1.6 $\pm$ {\tiny 3.58}} &
    40.4 $\pm$ {\tiny 28.51} & 32.4 $\pm$ {\tiny 14.26} &
    30.0 $\pm$ {\tiny 13.87} & 34.0 $\pm$ {\tiny 14.75} &
    92.4 $\pm$ {\tiny 8.08} & \secondcell{98.0 $\pm$ {\tiny 2.00}} \\

    & BC-P &
    32.8 $\pm$ {\tiny 14.78} & 0.0 $\pm$ {\tiny 0.00} &
    33.2 $\pm$ {\tiny 19.65} & 38.0 $\pm$ {\tiny 12.57} &
    31.2 $\pm$ {\tiny 10.23} & 26.4 $\pm$ {\tiny 11.82} &
    83.2 $\pm$ {\tiny 15.53} & 87.2 $\pm$ {\tiny 2.28} \\
    
    & R-P &
    30.4 $\pm$ {\tiny 7.13} & 0.0 $\pm$ {\tiny 0.00} &
    0.0 $\pm$ {\tiny 0.00} & 28.4 $\pm$ {\tiny 10.79} &
    24.0 $\pm$ {\tiny 2.68} & 44.4 $\pm$ {\tiny 12.73} &
    62.4 $\pm$ {\tiny 15.36} & 50.4 $\pm$ {\tiny 14.41} \\

    & D‑REX &
    22.4 $\pm$ {\tiny 10.32} & 0.8 $\pm$ {\tiny 1.79} &
    \secondcell{48.0 $\pm$ {\tiny 13.83}} & \bestcell{58.0 $\pm$ {\tiny 11.40}} &
    \secondcell{40.4 $\pm$ {\tiny 14.03}} & \secondcell{45.2 $\pm$ {\tiny 14.37}} &
    \secondcell{96.4 $\pm$ {\tiny 4.10}} & 66.0 $\pm$ {\tiny 19.41} \\

    \cmidrule(lr){2-10}
    & \textbf{SPW (ours)} &
    \bestcell{45.2 $\pm$ {\tiny 15.38}} & \bestcell{46.8 $\pm$ {\tiny 14.86}} &
    \bestcell{50.0 $\pm$ {\tiny 16.46}} & \secondcell{57.6 $\pm$ {\tiny 12.60}} &
    \bestcell{53.2 $\pm$ {\tiny 10.77}} & \bestcell{46.4 $\pm$ {\tiny 6.42}} &
    \bestcell{99.2 $\pm$ {\tiny 1.15}} & \bestcell{98.4 $\pm$ {\tiny 1.50}} \\
    \midrule

    \multirow{5}{*}{200} &
    MR &
    \secondcell{28.4 $\pm$ {\tiny 13.02}} & 2.0 $\pm$ {\tiny 3.46} &
    23.2 $\pm$ {\tiny 13.23} & 24.8 $\pm$ {\tiny 10.95} &
    27.4 $\pm$ {\tiny 10.59} & \secondcell{44.0 $\pm$ {\tiny 13.04}} &
    \secondcell{94.8 $\pm$ {\tiny 6.57}} & 95.2 $\pm$ {\tiny 3.63} \\

    & BC-P &
    24.8 $\pm$ {\tiny 10.98} & 2.0 $\pm$ {\tiny 3.46} &
    19.2 $\pm$ {\tiny 14.61} & 29.2 $\pm$ {\tiny 13.48} &
    27.6 $\pm$ {\tiny 11.64} & 39.6 $\pm$ {\tiny 11.13} &
    93.6 $\pm$ {\tiny 14.31} & \secondcell{95.6 $\pm$ {\tiny 8.76}} \\

    & R-P &
    23.2 $\pm$ {\tiny 7.82} & 0.4 $\pm$ {\tiny 0.89} &
    0.0 $\pm$ {\tiny 0.00} & 28.0 $\pm$ {\tiny 9.21} &
    20.0 $\pm$ {\tiny 3.63} & 38.0 $\pm$ {\tiny 12.58} &
    82.4 $\pm$ {\tiny 7.27} & 55.6 $\pm$ {\tiny 14.79} \\

    & D‑REX &
    24.0 $\pm$ {\tiny 14.66} & \secondcell{11.6 $\pm$ {\tiny 13.59}} &
    \secondcell{39.2 $\pm$ {\tiny 26.12}} & \secondcell{43.2 $\pm$ {\tiny 12.74}} &
    \secondcell{46.4 $\pm$ {\tiny 8.76}} & 42.4 $\pm$ {\tiny 12.03} &
    93.6 $\pm$ {\tiny 8.88} & 58.8 $\pm$ {\tiny 12.83} \\

    \cmidrule(lr){2-10}
    & \textbf{SPW (ours)} &
    \bestcell{63.2 $\pm$ {\tiny 14.60}} & \bestcell{41.2 $\pm$ {\tiny 15.45}} &
    \bestcell{74.8 $\pm$ {\tiny 11.64}} & \bestcell{64.4 $\pm$ {\tiny 16.52}} &
    \bestcell{59.0 $\pm$ {\tiny 6.00}} & \bestcell{48.4 $\pm$ {\tiny 5.90}} &
    \bestcell{100.0 $\pm$ {\tiny 0.00}} & \bestcell{98.4 $\pm$ {\tiny 2.61}} \\
    \bottomrule
  \end{tabular}
  \caption{Success rate (\%) comparison of methods that integrate demonstrations and preferences in eight Meta‑World tasks with 100 and 200 preference feedbacks.}
  \label{tab:dempref}
\end{table*}

\section{Experiments}
In this section, we conduct a series of experiments to investigate the following questions:
(1) Relative to state-of-the-art offline PbRL methods, does SPW outperform them?
(2) Compared with previous methods that combine demonstrations and preferences, does SPW offer better performance?
(3) After measuring similarity to expert demonstrations, is preference learning still necessary for high-quality reward learning?
(4) Can SPW effectively solve the credit assignment problem in preference learning?

\subsection{Settings}
Many offline PbRL methods are evaluated on D4RL~\citep{fu2020d4rl}, but it has been shown that high returns can be achieved even with incorrect rewards~\citep{shin2023benchmarks}, making it less suitable for evaluating reward learning.
We instead conduct experiments on Meta-World~\citep{yu2020meta}, a robotic manipulation benchmark widely used in PbRL research.  
We use the \textit{medium-replay} datasets provided by \citep{choi2024listwise}, which are collected from the replay buffers of online RL algorithms such as SAC \citep{haarnoja2018sac}. Expert demonstrations are generated by the SAC-trained policy, and preference segments are randomly sampled from the dataset. We also conducted experiments in the DeepMind Control Suite (DMControl)~\cite{tassa2018deepmind}. Details and results are provided in the Appendix. As our focus is on comparing reward learning methods, we fix the downstream algorithm and hyperparameters, using IQL~\citep{kostrikov2021offline} for all baselines to ensure fairness. Detailed parameter settings for each baseline are provided in the Appendix. To evaluate the effectiveness of combining the two types of feedback, our experiments are conducted using only one expert demonstration and a limited amount of preference feedback, specifically 100 or 200 labels. The results demonstrate that even under such minimal human supervision, our method achieves strong performance.

\subsection{Comparison with Offline PbRL Baselines}
To answer research question (1), we compare SPW with five offline PbRL baselines.
 Markovian Reward (MR), Preference Transformer (PT)~\cite{kim2023preference}, Offline PbRL (OPRL)~\cite{shin2023benchmarks}, Inverse Preference Learning (IPL)~\cite{hejna2024ipl}, and Listwise Reward Estimation (LiRE)~\cite{choi2024listwise}. The experimental results are shown in Table~\ref{tab:pbrl}.
MR method estimates reward with a simple Markovian MLP and performs poorly in most environments, reflecting its limited expressiveness.
OPRL leverages model ensembles for active query selection, but yields only marginal improvements over MR.
PT replaces MLP with a transformer to capture temporal dependencies, improving tasks such as \textit{peg-unplug-side}, \textit{sweep-into}, but it underperforms on others, probably due to increased model complexity.
IPL bypasses reward modeling and directly learns policies from preferences, but consistently lags behind, underscoring the challenges of this strategy.
LiRE reformulates preferences as listwise rankings and replaces the BT model with a linear model, but still falls short of SPW.

Although SPW uses an extra expert demonstration, it builds on the simplest PbRL backbone (MR) without adding additional architectures, sampling strategies, or ranking mechanisms, and still outperforms all existing offline PbRL baselines. This shows that even minimal expert input can substantially improve preference learning. 
The complete learning curves for each method are provided in the Appendix.
In the next section, we compare SPW with methods that combine demonstrations and preferences to further highlight its advantages.

\subsection{Evaluating Demo‑Preference Integration}
\subsubsection{Comparing to Demo–Preference Baselines}
Prior work on combining preferences and demonstrations largely focuses on online settings. In those settings, human-in-the-loop experiments require real-time interaction. Here, we instead evaluate in the offline setting, which offers a practical and reproducible framework for comparing different approaches.

To answer research question (2), we adapt several representative algorithms to the offline setting for fair comparison. For the preference learning component, we use the standard offline PbRL algorithm MR across all methods.

Inspired by \citep{ibarz2018reward}, we first initialize the policy with behavior cloning (BC) from expert demonstrations and then apply PbRL using MR. We refer to this method as BC Pretraining (BC-P).  We follow the D-REX~\cite{brown2019drex} procedure to generate synthetic preference pairs between expert and general trajectories. We denote this method as D-REX. We also implement a method based on \citep{biyik2021bayesian}. In this method, a reward model is first pre-trained using the reward signal computed by SEABO~
\cite{lyu2024seabo} as supervision. It is then refined using preference feedback. We refer to this method as Reward Pretraining (R-P).

The results, shown in Table~\ref{tab:dempref}, indicate that SPW outperforms all other methods in most tasks. Neither policy pretraining nor reward pretraining yields significant benefits. Although pretraining with behavior cloning or expert-based reward models is common, these methods strictly follow the expert distribution \citep{ross2011dagger,lee2021pebble}, causing the network to overfit and ultimately limiting downstream performance. These observations highlight the limitations of sequentially applying demonstrations followed by preferences.

The D-REX–based data augmentation method achieves moderate improvements by enriching the preference dataset. However, it still falls short of SPW. In particular, we observe that D-REX tends to degrade performance when the number of preferences increases to 200. This is probably due to conflicts between augmented preferences (from expert–general trajectory pairs) and true preferences. These conflicts ultimately lead to a less accurate reward model. In contrast, SPW consistently improves as more preference data become available. This demonstrates better sample efficiency and robustness across feedback scales.

\subsubsection{Necessity of Preferences After Using Demonstrations}
To answer question (3), we also carried out experiments using only expert demonstrations. Specifically, we evaluate behavior cloning (BC) trained solely on expert trajectories, as well as SEABO, which assigns rewards based on the negative exponentiated distance between agent and expert transitions. As shown in Table~\ref{tab:dempref}, both approaches perform poorly on Meta-World. This indicates that relying solely on a small amount of expert data, whether through supervised imitation or similarity-based reward learning, fails to achieve high success rates.

\begin{figure}[t]
    \centering
    \includegraphics[width=\linewidth]{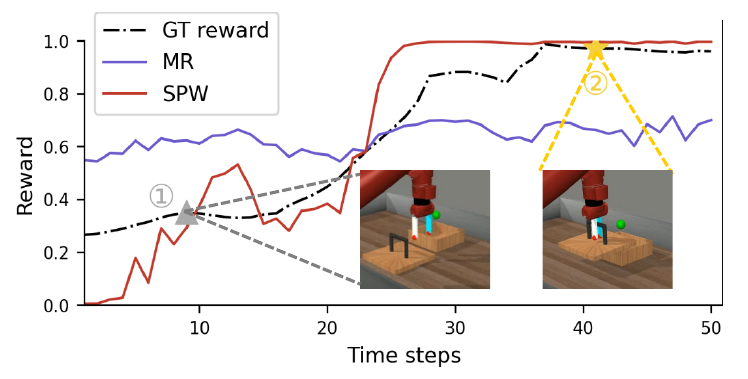}
    \caption{Normalized reward profiles of MR, SPW, and GT within a trajectory segment in the \textit{box-close} task. Snapshots from selected positions along the segment are shown for visual
reference.}
    \label{fig:credit}
\end{figure}

\begin{figure}[t]
    \centering
    \includegraphics[width=\linewidth]{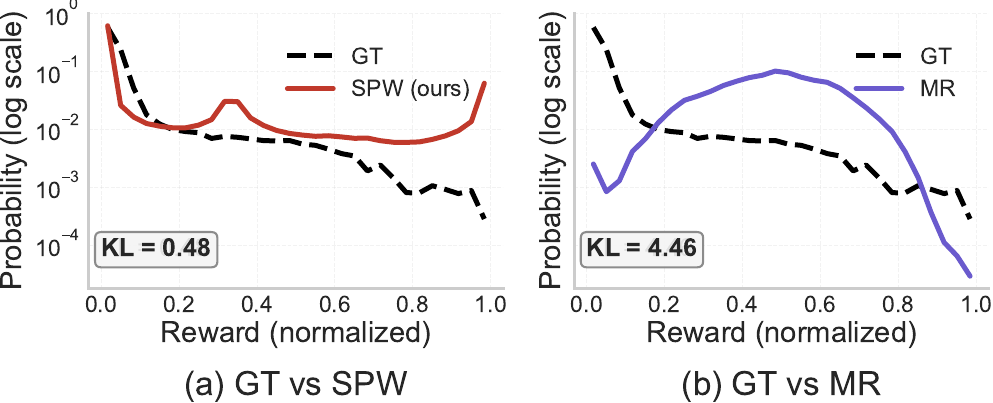}
    \caption{Comparison of the reward distributions learned by MR and SPW with the ground‑truth (GT) reward in \textit{peg‑unplug‑side} task.    
    The plot annotates the KL divergence between each learned distribution (SPW and MR) and the GT distribution.}
    \label{fig:dists}
\end{figure}

\subsection{Reward Analysis}
\subsubsection{Reward Distribution}
Figure~\ref{fig:dists} illustrates how MR and SPW distribute the reward probability mass in the \textit{peg-unplug-side} task.
The ground-truth (GT) reward histogram is sharply right-skewed. More than
40\,\% of the transitions receive a reward below 0.05.
SPW follows this long right-tailed pattern closely. However, it
assigns more probability to high rewards than GT. This suggests that it rewards a somewhat broader set of success-related
states. MR, by contrast, places the bulk of its mass around
0.4--0.5. In other words, MR treats most states as
“average,” and SPW, like GT, clearly separates commonplace transitions
from truly successful ones.

The KL numbers underscore this qualitative picture.
\(\mathrm{KL}(\text{GT}\!\parallel\!\text{SPW}) = 0.48\) is almost an
order of magnitude smaller than
\(\mathrm{KL}(\text{GT}\!\parallel\!\text{MR}) = 4.46\).
Thus, MR collapses the distribution toward the middle, while SPW captures the essential skew of the real reward signal far more faithfully.

\subsubsection{Reward Credit Assignment}
To answer research question (4), we examine whether SPW achieves effective credit assignment within a trajectory segment by analyzing per-transition rewards. Figure~\ref{fig:credit} shows results for the \textit{box-close} and \textit{dial-turn} tasks. SPW’s reward profile closely follows the ground-truth (GT) reward and produces a highly differentiated distribution within each segment that highlights critical states. In contrast, MR assigns nearly uniform rewards across the entire segment and fails to highlight the important transitions.

This effect stems from SPW’s design: it re‑weights each transition according to its similarity to expert behavior. Transitions that matter receive larger gradients, while irrelevant ones are down‑weighted. This weighting sharpens the score landscape and produces a more discriminative reward distribution, enabling the agent to distinguish informative states from uninformative ones.

\subsection{Ablation Study}
We present the results of two ablation studies here, and provide further ablation results in the appendix.

\subsubsection{Effect of Temperature $\tau$ in Softmax}
\begin{figure}[t]
    \centering
    \includegraphics[width=0.95\linewidth]{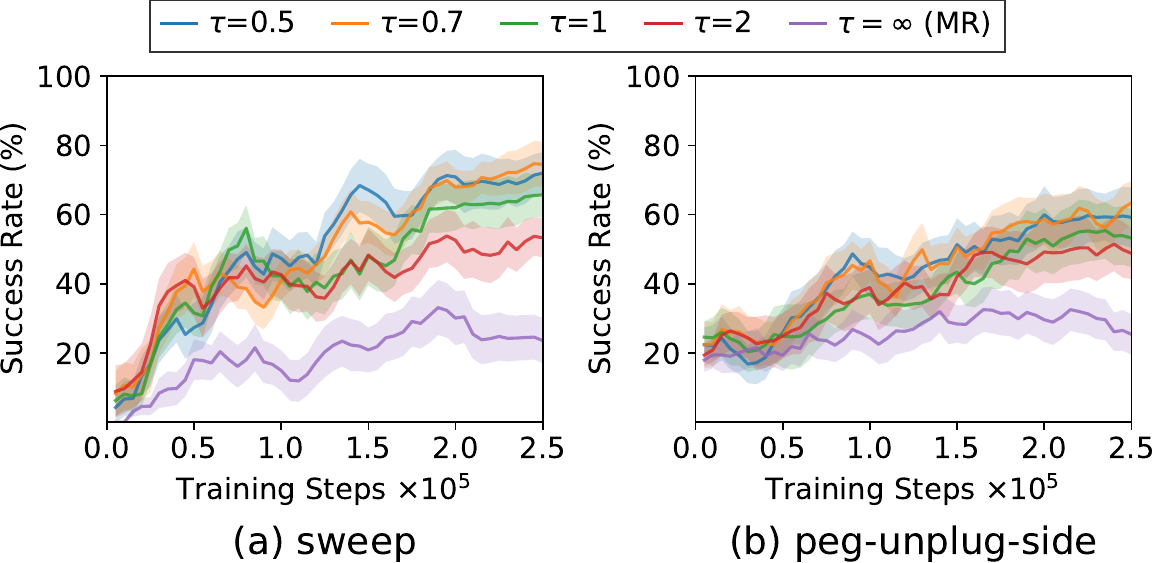}
    \caption{Average success rates of SPW when adjusting the temperature $\tau$. We use a total of 200 preference feedbacks.}
    \label{fig:tau}
\end{figure}

This section investigates the influence of the softmax temperature $\tau$ on policy success rate. Figure~\ref{fig:tau} reports performance when $\tau \in \{0.5, 0.7, 1, 2, \infty\}$, while the total number of preference feedbacks is fixed at 200. 

Larger $\tau$ values produce smoother weight distributions. When $\tau \to \infty$, the weights become uniform and our method degenerates to MR. The best performance is obtained at $\tau = 0.7$. As $\tau$ increases, the impact of the importance weights diminishes, leading to poorer credit assignment within trajectories and degraded results. When $\tau = \infty$, no weighting is applied, and the outcome is clearly worse than the weighted variants. 

This further confirms that our approach improves reward credit assignment through preference-trajectory weighting.

\subsubsection{Weight Integration Variant}
After computing the weights between preference--trajectory transitions and expert trajectories, we implement the redistribution strategy from Hindsight PRIOR~\citep{verma2024hindsight}.
Concretely, we first compute the per-step predicted rewards $\hat r_\psi(s_t, a_t)$ for each trajectory segment, then sum them to obtain the segment-level reward: $R_\psi(\sigma) = \sum_{t \in \sigma} \hat r_\psi(s_t,a_t).$
This total reward is redistributed to each step according to its learned weight $w_t$.
The network is trained with an additional mean-squared-error (MSE) loss.
Each step’s target reward becomes its weight $w_t$ times the segment return.
We denote this weighted-redistribution variant as RD.
The prior loss is defined as:
\[
\mathcal{L}_{\mathrm{prior}}
= \mathbb{E}_{\sigma\sim\mathcal{D}_{pref}}\left[ \sum_{t\in\sigma} \big(\hat r_\psi(s_t,a_t) - w_t\,R_\psi(\sigma)\big)^2 \right].
\]

As shown in Table~\ref{tab:redistribution}, our method (SPW) consistently outperforms RD. 
We attribute the gap to the extra MSE loss introduced during redistribution. 
This extra loss can conflict with the original preference loss. 
By contrast, SPW integrates the weights with a simple weighted sum, without introducing extra objectives that could interfere with training.

\begin{table}[t]
  \centering
  \footnotesize
  \setlength{\tabcolsep}{4pt} 
  \begin{tabular}{c|l|cccc}
    \toprule
    \textbf{Feedbacks} & \textbf{Algo.} &
    sweep & handle‑pull & dial-turn \\
    \midrule \midrule
    \multirow{2}{*}{100} &
    RD &
    39.2 $\pm$ {\tiny 18.68} &
    31.2 $\pm$ {\tiny 10.23} &
    34.4 $\pm$ {\tiny 13.30} \\

    & SPW &
    \textbf{50.0 $\pm$ {\tiny 16.46}} &
    \textbf{50.4 $\pm$ {\tiny 10.77}} &
    \textbf{46.4 $\pm$ {\tiny 6.42}} \\
    \midrule

    \multirow{2}{*}{200} &
    RD & 
    36.8 $\pm$ {\tiny 12.38} &
    32.8 $\pm$ {\tiny 12.92} &
    38.4 $\pm$ {\tiny 11.05} & \\

    & SPW & 
    \textbf{74.8 $\pm$ {\tiny 11.64}} &
    \textbf{59.0 $\pm$ {\tiny 6.00}} &
    \textbf{48.4 $\pm$ {\tiny 5.90}} \\
    \bottomrule
  \end{tabular}
  \caption{Comparison between RD and SPW in three Meta‑World tasks with 100 and 200 preference feedbacks. }
  \label{tab:redistribution}
\end{table}

\begin{figure}[t]
    \centering
    \includegraphics[width=0.95\linewidth]{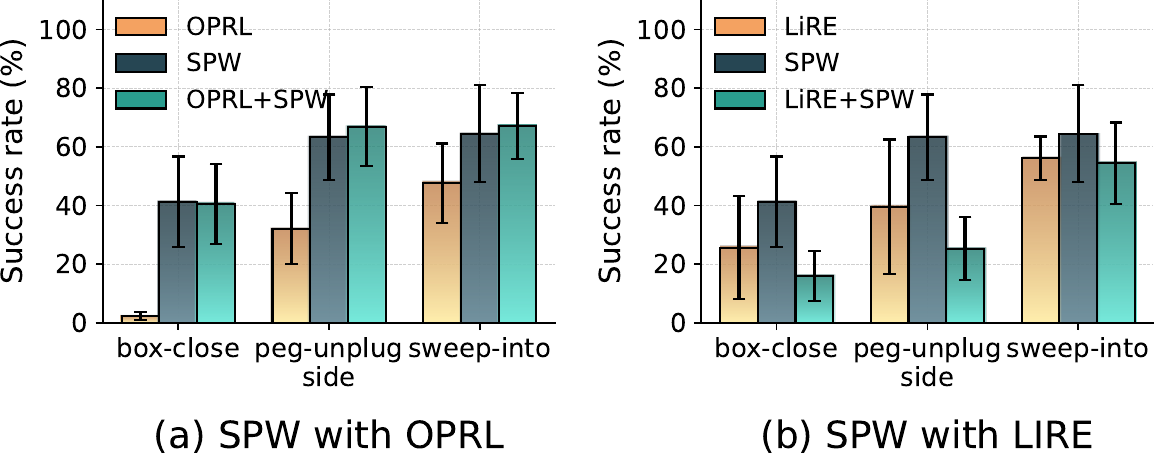}
    \caption{Combining SPW with other PbRL baselines with 200 preference feedbacks.}
    \label{fig:cwob}
\end{figure}

\section{Additional Analyses of SPW}
\subsection{Compatibility with Other Methods}
To investigate the compatibility of SPW with other PbRL techniques, we evaluated SPW combined separately with OPRL and with LiRE, using 200 preference queries. 
For the OPRL+SPW setting, we actively sampled new preference pairs based on the disagreement between the reward model. 
We then recomputed the weights for each newly acquired pair. 
The results show that in \textit{peg‑unplug‑side} and \textit{sweep‑into} tasks, fusing the two methods produces a synergistic boost over the baseline (MR). 
However, in \textit{box-close}, the combined variant performs roughly the same as vanilla SPW.
It appears to be limited by OPRL's own ceiling in that task. 

When we combined SPW with LiRE, the performance deteriorated instead. 
A likely reason is that LiRE employs a linear preference model rather than the exponential Bradley‑Terry model. 
Because it lacks an explicit constraint on reward‑difference magnitudes, our weighting scheme can cause the reward gap between preferred segments to explode. 
This ultimately hurts learning and leads to results worse than either method alone.

\subsection{Human Experiments}
Table~\ref{tab:human} reports results from training reward functions on real human preference feedbacks. 
We collected 200 preference labels in \textit{sweep-into} and \textit{box-close} tasks by asking experienced annotators to compare segments rendered from policies of different levels.
We then compared our method against PT and D-REX. 
The results show that even under real human supervision, our approach achieves strong performance. 
This is likely because human preferences over trajectories are driven by a few key states. 
SPW can identify and leverage these states effectively. 
These findings suggest that SPW holds strong potential in settings involving genuine human feedback.

\begin{table}[t]
  \centering
  \footnotesize
  \setlength{\tabcolsep}{4pt} 
  \begin{tabular}{c|cccc}
    \toprule 
    \textbf{Task} &
    PT & D-REX & SPW \\
    \midrule \midrule
    box-close &
    19.2 $\pm$ {\scriptsize 9.47} &
    2.0 $\pm$ {\scriptsize 3.46} &
    \textbf{56.4 $\pm$ {\scriptsize 8.71}}\\

    sweep-into &
    39.2 $\pm$ {\scriptsize 15.81} &
    40.4 $\pm$ {\scriptsize 9.48} &
    \textbf{70.0 $\pm$ {\scriptsize 9.70}} & \\
    \bottomrule
  \end{tabular}
  \caption{Average success rates with 200 real human feedbacks}
  \label{tab:human}
\end{table}

\section{Conclusion}
In this paper, we introduced SPW, a lightweight method that jointly learns from demonstrations and preferences. Unlike prior approaches, SPW unifies both feedback sources into a single‑stage learning framework without introducing additional loss terms, multi‑stage pipelines, or any online interaction. By extracting prior weights from a small set of expert demonstrations, SPW tackles the long‑standing credit‑assignment challenge in preference‑based reward learning. Extensive experiments and ablations show that, even with minimal human feedback, SPW delivers substantial performance gains over both pure PbRL and hybrid baselines. Our reward analysis further confirms its effectiveness in assigning credit accurately across trajectories. 


\bibliography{aaai2026}

\end{document}